\documentclass[sigconf]{acmart}

\copyrightyear{2025}
\acmYear{2025}
\setcopyright{rightsretained}
\acmConference[HAI '25]{Proceedings of the 13th International Conference on Human-Agent Interaction}{November 10--13, 2025}{Yokohama, Japan}
\acmBooktitle{Proceedings of the 13th International Conference on Human-Agent Interaction (HAI '25), November 10--13, 2025, Yokohama, Japan}
\acmDOI{10.1145/3765766.3765827}
\acmISBN{979-8-4007-2178-6/25/11}

\usepackage{algorithmic,algorithm}
\usepackage{tikz}
\usepackage{fontawesome5}
\usepackage{color}
\usepackage[dvipsnames, svgnames,x11names]{xcolor}
\usepackage{subcaption}

\begin{document}

\title{Towards Human Engagement with Realistic AI Combat Pilots}

\author{Ardian Selmonaj}
\affiliation{%
  \institution{Istituto Dalle Molle di Studi sull'Intelligenza Artificiale}
  \city{Lugano}
  \country{Switzerland}}
\email{ardian.selmonaj@idsia.ch}

\author{Giacomo Del Rio}
\affiliation{%
  \institution{Istituto Dalle Molle di Studi sull'Intelligenza Artificiale}
  \city{Lugano}
  \country{Switzerland}}
\email{giacomo.delrio@idsia.ch}

\author{Adrian Schneider}
\affiliation{%
 \institution{Armasuisse Science+Technology}
 \city{Thun}
 \country{Switzerland}}
\email{adrian.schneider@armasuisse.ch}

\author{Alessandro Antonucci}
\affiliation{%
  \institution{Istituto Dalle Molle di Studi sull'Intelligenza Artificiale}
  \city{Lugano}
  \country{Switzerland}}
\email{alessandro.antonucci@idsia.ch}

\renewcommand{\shortauthors}{Selmonaj et al.}

\begin{abstract}
   We present a system that enables real-time interaction between human users and agents trained to control fighter jets in simulated 3D air combat scenarios. The agents are trained in a dedicated environment using \emph{Multi-Agent Reinforcement Learning}. A communication link is developed to allow seamless deployment of trained agents into \emph{VR-Forces}, a widely used defense simulation tool for realistic tactical scenarios. This integration allows mixed simulations where human-controlled entities engage with intelligent agents exhibiting distinct combat behaviors. Our interaction model creates new opportunities for human-agent teaming, immersive training, and the exploration of innovative tactics in defense contexts.
  
\end{abstract}



\received{18 July 2025}
\received[accepted]{12 August 2025}

\maketitle

\section{Introduction}
Extending the immense potential of \emph{Artificial Intelligence}~(AI) from strategic tasks to safety-critical domains such as air combat requires realistic simulations and interactive setups that enable humans to influence and evaluate AI agents during training and execution. While several combat environments exist~\cite{JSBsim_envs, aircombat_survey}, only few offer realistic flight dynamics, and even fewer enable direct interaction between AI models and human users~\cite{rl_interaction1, rl_interaction2, rl_interaction3}. Lockheed Martin's pioneering work at DARPA’s AlphaDogfight Trials first emphasized the significance of human-AI interactions in 1-vs-1 air combat scenarios~\cite{darpa}. Incorporating humans in the loop is essential to ensure AI systems remain aligned with human judgment and safety constraints, while also fostering trust and leveraging complementary strengths in high-risk domains. However, human-AI interaction in multi-agent defense setups remains relatively unexplored.

\begin{figure}[htb!]
  \centering
  \begin{subfigure}[b]{0.27\columnwidth}
    \includegraphics[width=\linewidth]{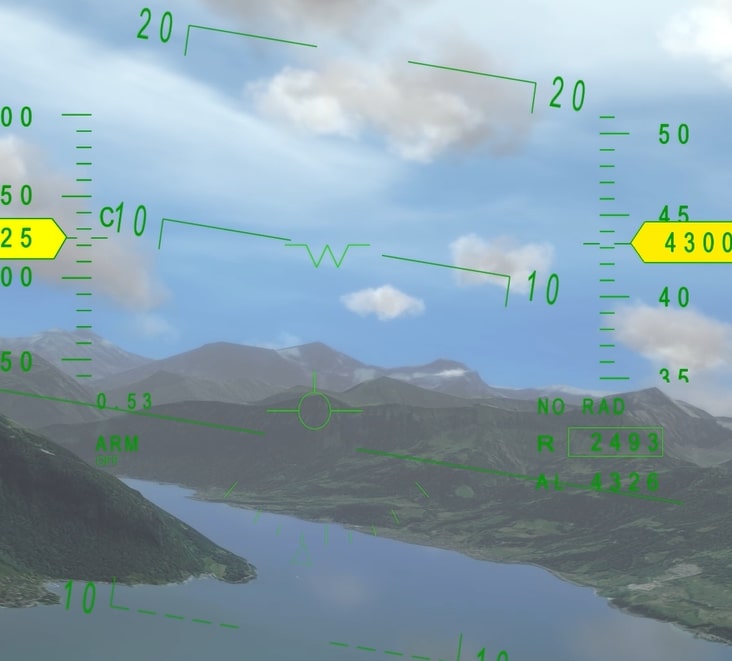}
    \caption{FPV.}
    \label{fig:vrf-fpv}
  \end{subfigure}
  \begin{subfigure}[b]{0.405\columnwidth}
    \includegraphics[width=\linewidth]{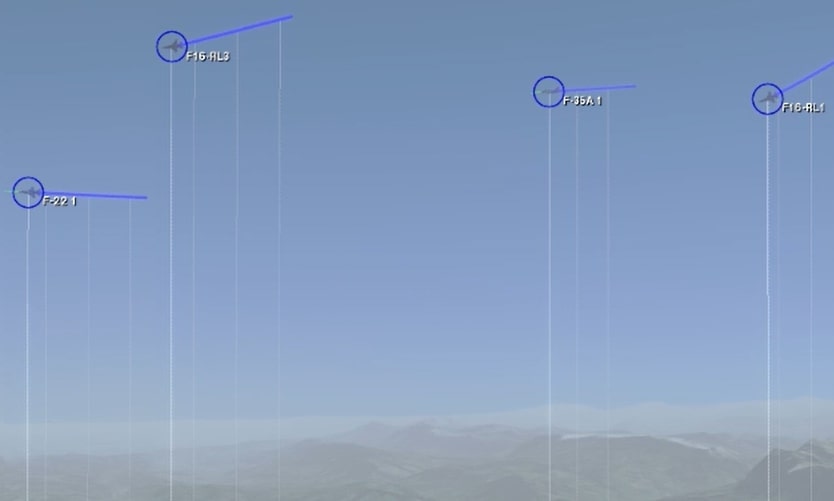}
    \caption{2-vs-2 scenario.}
    \label{fig:vrf-multi}
  \end{subfigure}
  \begin{subfigure}[b]{0.29\columnwidth}
    \includegraphics[width=\linewidth]{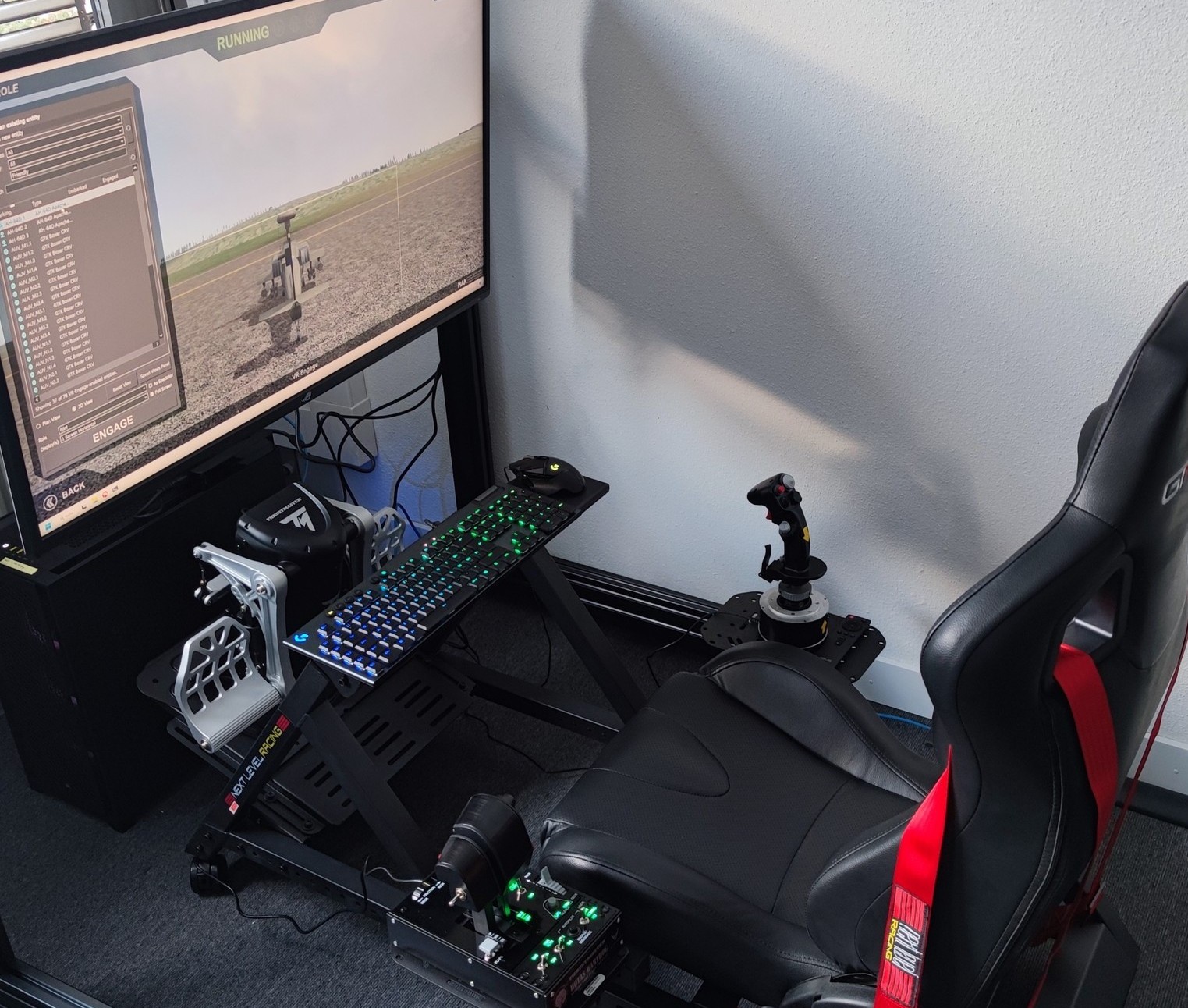}
    \caption{Cockpit.}
    \label{fig:vrf-cockpit}
  \end{subfigure}
  \caption{Scenes of VR-Forces as (a) \emph{First Person View}~(FPV), (b) combat situation, and (c) simulator cockpit.}
  \Description[Components in our approach.]{Scenarios of simulation tool named VR-Forces and Cockpit simulation component used in our approach.}
  \label{fig:vrf-scenes}
\end{figure}

This work contributes a modular framework to bridge this gap. Our system uses \emph{Multi-Agent Reinforcement Learning}~(MARL) to train agents within a custom-built 3D environment ensuring accurate flight dynamics simulation. Using the IEEE standard \emph{Distributed Interactive Simulation}~(DIS)~\cite{disoriginal}, we developed a communication interface allowing seamless deployment of MARL agents into \emph{VR-Forces}\footnote{\hyperlink{https://www.mak.com/mak-one/apps/vr-forces}{mak.com/vr-forces}.}~(VR-F), a tactical simulator used by defense organizations (Figs.~\ref{fig:vrf-fpv} and~\ref{fig:vrf-multi}). This integration enables real-time interactions between human-operated entities and AI-controlled aircraft, fostering competitive agent behavior and enhanced training realism for military personnel. To the best of our knowledge, this is the first approach to integrate MARL agents into VR-F for human-agent teaming. This opens new pathways for tactical innovation and exploration of imaginative strategies in operational tasks.

\section{Reinforcement Learning Agent Model}
MARL~\cite{marl_book} involves $n$ agents learning in an environment, modeled as a \emph{Markov Game}~\cite{markov_game_def} with the tuple $(\mathcal{N}, \mathcal{S}, \{\mathcal{A}^i\}_{i=1}^n, P, \rho, \{R^i\}_{i=1}^n, \gamma)$. Here, $\mathcal{N} = \{1, \dots, n\}$ is the agent set, $\mathcal{S}$ the state space, $\mathcal{A}^i$ the action space of agent $i \in \mathcal{N}$, and initial state $s_0 \sim \rho$. Given $s_t$, an agent selects action $a_t^i \in \mathcal{A}^i$ via its policy $\pi^i(a \mid s_t)$, forming joint action $\boldsymbol{a}_t=(a_t^1,\dots,a_t^n)$. The environment transitions to $s_{t+1} \sim P(\cdot \mid s_t,\boldsymbol{a}_t)$, and agent $i$ receives reward $R^i(s_t,\boldsymbol{a}_t,s_{t+1})$. Each agent maximizes its expected return $\mathbb{E}_{\pi^i}[\sum_{t=0}^\infty \gamma^t R^i(s_t,\boldsymbol{a}_t,s_{t+1})]$, with $\gamma \in [0,1)$. We train our agents using \emph{(i)}~MA-PPO~\cite{mappo}, \emph{(ii)}~Actor-Critic network~\cite{actor_critic}, \emph{(iii)}~CTDE paradigm~\cite{ctde}, with global information during training while allowing independent agent execution in any $m$-vs-$n$ combat scenario, \emph{(iv)}~curriculum learning, which gradually increases task difficulty, and \emph{(v)}~self-play, where agents learn by competing against themselves. To promote strategic diversity, agents are trained for three distinct \emph{control policies}: \emph{(i)~Attack $\pi_a$}, for aggressively destroying opponents; \emph{(ii)~Engage $\pi_e$}, seeking a positional advantage behind enemy aircraft; and \emph{(iii)~Defend $\pi_d$}, aiming to flee from opponents. Each agent further learns a commander policy $\pi_c$ that decides which control policy to activate. When agents and enemies use the same control policies, with enemies adopting a mixed strategy with probability rates $\pi_a$ ($40\text{\footnotesize{\%}}$), \, $\pi_e$ ($40\text{\footnotesize{\%}}$), and $\pi_d$ ($20\text{\footnotesize{\%}}$), agents with $\pi_c$ achieve an $80 \text{\footnotesize{\%}}$ win rate in $10 \text{-vs-} 10$ air combats.

Modeling realistic and accurate aircraft physics is crucial to achieve strong combat performance. Therefore, we developed a custom training environment based on \emph{JSBSim}~\cite{jsbsim}, an open-source flight dynamics simulator grounded in real-world physics, which runs at integration frequency of $100$ Hz.  While we use an F-16 model, our environment supports other aircraft types as well.

\section{Interaction Process}

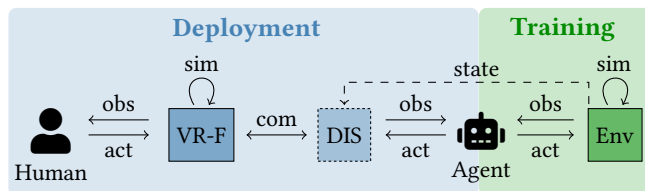
\begin{figure}[htb!]
    \centering

    \begin{tikzpicture}
        \node[rectangle, draw=none, fill=SteelBlue!20, minimum width=6.25cm, minimum height=2.5cm, rounded corners=3pt] at (0.56,0.44) {};
        \node at (0.6, 1.4) {\textcolor{SteelBlue}{\textbf{\large Deployment}}};
        
        \node[rectangle, draw=none, fill=Green4!20, minimum width=2.3cm, minimum height=2.5cm, rounded corners=3pt] at (4.84,0.44) {};
        \node at (4.8, 1.4) {\textcolor{Green4}{\textbf{\large Training}}};
        
        \node at (-2,0.05) {\Huge \faUser};
        \node at (3.72,0.05) {\huge \faRobot};

        
        \node[draw, fill=SteelBlue!70, minimum width=0.7cm, minimum height=0.7cm] at (0,0) {VR-F};
        \draw[->] (-0.1,0.4) to[out=120, in=60, looseness=7] (0.1,0.4) node[pos=0, yshift=27pt] {sim};

        \node[draw, fill=SteelBlue!50, dashed, dash pattern=on 1pt off 1pt, minimum width=0.7cm, minimum height=0.7cm] at (1.9,0) {DIS};

        \node at (-2, -0.5) {Human};
        \node at (3.72, -0.5) {Agent};

        \draw[->] (-1.5, 0.0) -- (-0.7,0.0) node[midway, below] {act};
        \draw[->] (-0.7, 0.2) -- (-1.5, 0.2) node[midway, above] {obs};

        \draw[<->] (0.6, 0.0) -- (1.4, 0.0) node[midway, above] {com};

        \draw[<-] (2.45, 0.0) -- (3.25,0.0) node[midway, below] {act};
        \draw[<-] (3.25, 0.2) -- (2.45, 0.2) node[midway, above] {obs};

        \draw[->] (4.2, 0.0) -- (4.95,0.0) node[midway, below] {act};
        \draw[->] (4.95, 0.2) -- (4.2, 0.2) node[midway, above] {obs};
        \draw[->, dashed] (5.15,0.2) -- ++(0,0.5) -- ++(-3.25,0) -- ++(0,-0.3) node[pos=0, xshift=50pt, yshift=5pt] {state};

        \node[draw, fill=Green4!60, minimum width=0.7cm, minimum height=0.7cm] at (5.5,0) {Env};
        \draw[->] (5.4,0.4) to[out=120, in=60, looseness=7] (5.6,0.4) node[pos=0, xshift=156pt, yshift=27pt] {sim};

    \end{tikzpicture}
    
    \caption{In training, our custom environment (\emph{Env}) simulates (\emph{sim}) dynamics. Agents observe (\emph{obs}) states and respond with actions (\emph{act}). In deployment, DIS communicates (\emph{com}) to VR-F either JSBSim states (dashed) or actions for state simulation. Human users interact with VR-F via input devices.}
    \Description[Interaction process between humans and AI agents.]{Interaction diagram between involved components. Human interacts with VR-F directly, the agent is simulated in the environment and its data is transferred to VR-F by the established DIS connection.}
    \label{fig:interaction_process}
\end{figure}

Fig.~\ref{fig:interaction_process} shows the interaction process between human users and MARL agents. Training is done in our custom JSBSim-based environment, since VR-F does not support this. Afterward, agents are deployed in VR-F for higher realism and diverse tactical scenarios.

This integration is achieved through the real-time protocol DIS, which enables multiple simulators to synchronize entity information over a network for joint virtual environments. We employ the open-source version \emph{OpenDIS}\footnote{\hyperlink{https://github.com/open-dis/open-dis-python}{github.com/open-dis-python}.}, which operates by sending and receiving serialized \emph{Protocol Data Units}~(PDUs) via UDP/IP sockets. We implemented a DIS communication class (Algorithm~\ref{algo:dis_state_incoming}) that records PDU state events capturing entity location, orientation, and velocity. Upon receiving and processing a new state $s_t$, agents and enemies compute actions $a^i_t$ using their currently activated control policy $\pi^i_{k}(\cdot \mid s), k\in \{a,e,d\}, i\in \mathcal{N}$, determined by their commander $\pi_c$. Two types of DIS messages are supported for sending data to VR-F: \emph{(i)} simulate the agent’s dynamics locally in JSBSim and transmit the updated state to VR-F, as depicted by the dashed line from \emph{Env} to \emph{DIS} in Fig.~\ref{fig:interaction_process}, or \emph{(ii)} transmit the computed action $a^i_t$ to VR-F and let VR-F simulate the resulting state evolution.

Human users operate their entities via controllers or joysticks in our custom-built simulator cockpit (Fig.~\ref{fig:vrf-cockpit}). They can cooperate or compete with AI agents, and multiple humans can participate to form mixed human-AI teams. Since AI agents may reason differently, they can generate novel maneuvers valuable for training.

\begin{algorithm}[htb!]
\caption{DIS Connection and Message Passing}
\begin{algorithmic}[1]
\STATE Open UDP socket on given port and address
\STATE Start receiver thread
\WHILE{receiving}
    \STATE Get UDP packet and parse to PDU
    \IF{PDU is EntityStatePdu \textbf{and} matches VRF filter}
        \STATE Trigger state handling callback
    \ENDIF
    \IF{new agent state $s_t$ received}
        \STATE Get action $a_t^i \sim \pi^i(\cdot \mid s_t)$ for all agents $i \in \mathcal{N}$
        \STATE Send encoded PDU command to VR-F as UDP packet 
    \ENDIF
\ENDWHILE
\end{algorithmic}
\label{algo:dis_state_incoming}
\end{algorithm}

\section{Conclusions and Future Work}

This work introduces a novel integration of MARL agents into the complex and realistic wargaming simulation tool VR-F, enabling real-time interaction between human users and AI-controlled aircraft. Moving forward, we plan to enhance our MARL model with more advanced planning capabilities to capture complex skills and tactics. By collecting behavioral data during human-agent interactions and obtaining explicit feedback from real pilots, we aim to improve model realism by going beyond self-play to imitation learning and investigating hybrid algorithms that combine MARL with supervised learning techniques. This bidirectional training setup not only allows agents to become more competitive to potentially surpass human capabilities but also supports simultaneous skill improvement for military personnel. 

Future research will explore how air combat strategies evolve when purely driven by AI compared to human-AI collaboration, with a focus on imaginative trajectories that deviate from human expectations. This may uncover novel strategies that go beyond conventional human understanding. Additionally, as we make use of DIS, our communication setup is not limited to VR-F but also allows seamless connection to other simulation tools as well.

To further enhance training complexity and realism, we are developing a system that trains MARL agents directly within VR-F. This is achieved using a split architecture, where one part encapsulates the VR-F engine via its C++ API, and the other satisfies the requirements of standard Gymnasium\footnote{\hyperlink{https://gymnasium.farama.org/}{gymnasium.org.}} environments in Python. State messages are exchanged over TCP/IP using a dedicated binary protocol and multiple parallel environments can run with an iteration rate of about 10 Hz. 

Ethical aspects are integral to our work, as transparency and safety remain paramount in AI-driven air combat simulations. Our approach strives to maintain human oversight by keeping AI agents supportive rather than autonomous in life-critical scenarios, thereby fostering trust and responsible deployment in operational settings.

\bibliographystyle{ACM-Reference-Format}
\bibliography{refs}

\end{document}